\documentclass{article}
\usepackage{natbib}

\usepackage[T1]{fontenc}
\usepackage{tikz}
\usetikzlibrary[arrows.meta,bending]
\usetikzlibrary{positioning}
\usetikzlibrary{decorations.text}
\usetikzlibrary{circuits.logic.US,circuits.logic.IEC,fit}

\usepackage{textcomp}
\usepackage{graphicx}
\usepackage{setspace}
\usepackage{graphicx}
\usepackage{pifont}
\usepackage{geometry}
\usepackage{algorithm}
\usepackage{algorithmic}

\usepackage{pdflscape}
\usepackage{afterpage}

\usepackage{amsmath}
\usepackage{amsfonts}
\usepackage{amssymb}
\usepackage{kbordermatrix}

\usepackage{url}
\usepackage{rotating}

\usepackage{caption}
\usepackage[english]{babel}
\usepackage{array}

\usepackage{subcaption} 
\usepackage[colorlinks=TRUE, linkcolor=blue, urlcolor=blue, citecolor=blue]{hyperref}



\begin{document}

\title{A Bimodal Network Approach to Model Topic Dynamics} 
\author{
\large
\textsc{Luigi Di Caro $^{1,3}$, Marco Guerzoni $^{1,2}$,}\\ \textsc{Massimiliano Nuccio $^{1,2}$, Giovanni Siragusa $^{1,3}$}
\\[2mm] 
\normalsize {$^{1}$ Despina, Big Data Lab} \\
  \normalsize {$^{2}$ Department of Economics and Statistics "Cognetti de Martiis", University of Turin, Italy}\\
 \normalsize {$^{3}$ Department of Computer Science, University, of Turin, Italy} \\ 
\vspace{-5mm}
}

\date{}

\maketitle

\vspace{2cm}
\begin{abstract}
\noindent 
This paper presents an intertemporal bimodal network to analyze the evolution of the semantic content of a scientific field within the framework of topic modeling, namely using the Latent Dirichlet Allocation (LDA). The main contribution is the conceptualization of the topic dynamics and its formalization and codification into an algorithm. To benchmark the effectiveness of this approach, we propose  three indexes which track the transformation of topics over time, their rate of birth and death, and the novelty of their content. Applying the LDA, we test the algorithm both on a controlled experiment and on a corpus of several thousands of scientific papers over a period of more than 100 years which account for the history of the economic thought.

Keywords: topic modeling, LDA, bimodal network, topic dynamics, economic thought

\end{abstract}

\footnotetext[1]{We would like to thank JSTOR (\url{www.jstor.org}) for providing the data and  DESPINA -Big Data Lab (\url{www.despina.unito.it}) and the Department of Computer Science at the Univeristy of Turin for financial support.} 
\vspace{2cm}

\section{Introduction\label{sec:Intro}} 

A crucial issue in the philosophy of science consists in the understanding of the evolution of scientific paradigms within a discipline. Following \citet[p.10]{kuhn1970}, a scientific paradigm can be thought as the set of assumptions, legitimate theories, methods, and experiments both adequately new to attract a group of scholars, to build a contribution to a field and to open enough the exploration of different directions of research.

In the traditional view, as developed for hard and mature sciences, the evolution of scientific paradigm consists in "\textit{the successive transition from one paradigm to another via revolution}" \citep[p.12]{kuhn1970}. However, a scientific field is usually composed by several research paradigms either competing or addressing different issues,  and a revolution in one of those necessarily involves effects and readjustments in the entire discipline. Moreover, each new paradigm carries the legacy of the existing knowledge of past paradigms, which is often recombined into the new one.
This is especially true for social sciences, in which the identification of clear scientific paradigms in the sense of Kuhn is often blurred and it is probably more correct referring to "research traditions" \citep{laudan1978}.

However, whether you call paradigms or traditions, the existence of patterns of thoughts which are legitimate contributions to a theory is undeniable.
Thus, we can postulate that the evolution of knowledge in a scientific field is generated among a community of researchers which share a semantic area to define specific research issues, describe methodologies, and lay down results. Thus, the heterogeneity of the research tradition of a scientific field can be described with semantic analysis. The idea that some measure of words co-occurrence reveals an underlying epistemic pattern and, therefore, it can capture the essence of evolution in science is not a new one. Despite the difficulty in programming, the first attempts date back to the work of \citet{callon1983} and refined when the first open code have been made available a decade later \citep{vlieger2011content, leydesdorff2011semantic}.  

The challenge of classifying science on the basis of its semantic content has found a renewal with the diffusion of machine learning techniques and, in particular, in the subfield of unsupervised learning \citep{Leydesdorff2015}. 
Topic modeling includes a family of algorithms \citep{BleiNgJordan2003}, which are particularly performant in extracting information from large corpora of textual data by reducing dimensionality. This feature has been clearly recognised in mapping science \citep{Suominen2015} or news \citep{dimaggio2013}. 
\citet{AlghamdiAlfalqi2015} review four major methods of topic modeling, including Latent Semantic Analysis (LSA), Probabilistic LSA, Latent Dirichelet Allocation (LDA) and Correlated Topic Model (CTM). The LDA proposed in \citep{BleiNgJordan2003} is one of the most diffused approaches. LDA retrieves latent patterns in texts on the basis of a probabilistic Bayesian model, where each document is a mixture of latent topics described by a multinomial distribution of words. One of the major limitations of LDA lies on its inability to model and represent relationships among topics over time \citep{AlghamdiAlfalqi2015}.

In this paper, we address a major recurring issue in topic modeling, that is the topic dynamics, or, in other words, we test a method to track the transformation of topics over time. As stated by \citet{Blei2006}, LDA is a powerful approach to reduce dimensionality, but it assumes that documents in a corpus are exchangeable. On the contrary, articles and themes are sequentially organized and evolve over time. Therefore, it is not only relevant to develop a statistical model to determine the evolving topics from a corpus of a sequential collection of documents, but also to measure and describe the transformation of topics and their appearance and disappearance.

In the literature of information retrieval, the dynamics of topics has been faced with two approaches \citep{Heetal2009}: a discriminative one monitors a change in the distribution of words or in the mixture over documents, while a generative approach searches for general topics over the whole corpus and, then, it assigns the documents which belong to each topic \citep{bolelli2009,Heetal2009}. 

Specifically \citet{Blei2006} introduced Dynamic Topic Modeling (DTM), a class of generative models in which the per document topic distribution and per topic word distributions are generated from the same distributions in a previous time frame. This approach has been very influential since it imposes a connection between the sets of topics at different periods and allows to track the evolution of a single topic over time. 
 
DTM performs very well in capturing the evolution of a single topic. However, the evolution of knowledge is much more complicated that the change of relative importance of words within a topic, since it may involve also the creation of new topics, their mutual re-combinations and, eventually their possible demise. The major contribution of the paper is the conceptualization and formalization of the evolution of knowledge, conceived as different streams of semantic content which continuously appears and disappears, merges and splits. Thereby we propose an original method based on inter-temporal bimodal networks of topics  compute the key elements in the evolution of knowledge. 

Moreover, the ultimate goal of the paper is not to track in detail what happens within a single topic, but rather to develop indexes which can measure at the aggregate level some properties of the observed knowledge dynamics, such as an overall degree of novelty or the level of turbulence at specific time windows.

The paper is organized as follows: in the next section, we suggest a method to analytically conceptualize and measure different patterns of topics evolution. Section \ref{sec:algorithm} translates it into an algorithm which calculate some measures of merging, splitting and novelty of the topics generated by the LDA. In section \ref{sec:toy}, a simple simulation tests the robustness of the method on artificial data. Finally, in Section \ref{sec:results}, the same algorithm is applied to a large dataset of papers in economics: main results are presented and discussed by describing the evolution of the topics in the economic science in the past century.

\section{A Conceptualization of Knowledge Evolution\label{sec:concept}}

In this paper, we focus on the dynamic evolution of topics over time. With DTM, each topic $K_{t}$ is linked to $K_{t+1}$ creating a topics chain which spans the years covered by the documents. 
Specifically, \citet{Blei2006} maps each topic at time t-1 into a topic in t by chaining the per document topic distribution  $\alpha _{t}$ and the per topic word distribution, $\beta _{t,k}$ in a sate space model with a Gaussian noise:

\begin{equation}
\beta _{t,k}|\beta _{t-1,k}\sim \mathcal{N}(\beta_{t-1,k},\sigma ^{2}I)
\end{equation}

\begin{equation}
\alpha _{t}|\alpha _{t-1}\sim \mathcal{N}(\alpha_{t-1},\delta ^{2}I)
\end{equation}

This approach is highly performing to track incremental changes of the same topic but it does not focus on revealing neither birth nor death nor possible combinations of topics and it imposes a constant number of topics within the model.
On the contrary, we are interested to discover the structural change of topics in a corpus and to understand the underlying topic dynamics which explain it. Thereby, we do not focus on the evolution of the single topic. The inter-temporal link across topics is not a constraint in the estimation of the model as in the DTM, but it is introduced ex-post in the empirical analysis by looking at the similarities (co-occurrence of words) amongst topics generated by independent LDAs. More in detail, while DTM models sequences of compositional random variables by chaining Gaussian distributions (thus directly embodying topics dynamics in the model), our approach operates on single and static LDAs in order to track and measure such dynamics out of the model.

The evolution  of a topic structure of a corpus accumulating knowledge overtime takes place because of two main reasons. 
On the one hand, any epistemic community (say for instance journalists or scientists) can shift their intellectual interest to new issues and problems, which will result in different choices, frequencies and co-occurrence of words. On the other hand, language is subject to a constant evolution, in which new words, named entities, acronyms, etc. appear while other ones disappear due to an increasingly lesser use of them by the same community. We rule out this second scenario, by assuming that in the short time frame the language is fairly stable.

Under this assumption, when comparing the topics generated by a topic modeling exercise in two different, although adjacent, time windows, we should be able to capture the evolution of the scientific debate and highlight the birth, death and recombination of topics.
On the one extreme, we can find a situation in which knowledge does not evolve and thus topics are stable. On the other, we figure out the maximum of turbulence in which new topics emerge without any semantic relation with the incumbent ones. In the latter case, we may assume the death of past topics and the birth of new ones. In between the two ideal cases, we can also draw a continuum in which we can observe both deaths and births of topics. 
Finally, in a most interesting scenario, rather than observing stability or turbulence, knowledge may evolve recombining existing topics in both old and new ones. Table \ref{ideal} summarizes five typical patterns of knowledge evolution and their interpretation within a topic modeling framework.

\begin{table}[h!]
\centering
\caption{Topic modeling and typical patterns of knowledge evolution}
\label{ideal}
\begin{tabular}{|l|l|}
\hline
Stability & a topic A exists at time t and t+1                             \\ \hline
Birth     & the topic A at time t+1 has no antecedent at time t            \\ \hline
Death     & the topic A at time t disappears at time t+1                   \\ \hline
Merging & multiple topics at time t combine in a new topic A at time t+1 \\ \hline
Splitting & multiple topics at time t+1 share an antecedent at time t      \\ \hline
\end{tabular}
\end{table}

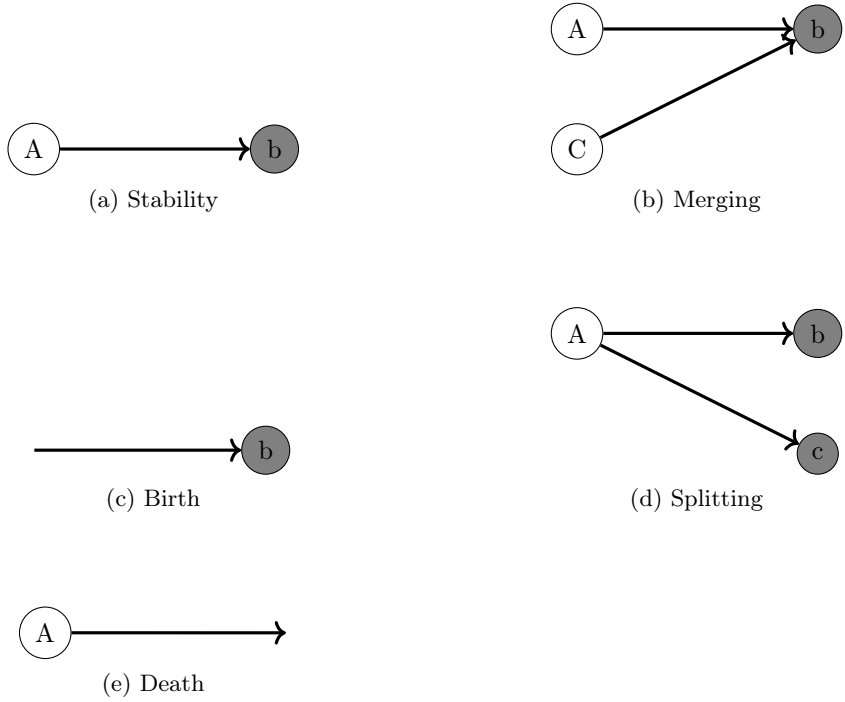
\begin{figure}[h]\caption{Ideal types of topic evolution}\label{types} 
\begin{tabular}{cc}
\begin{subfigure}[b]{0.45\textwidth}\hfill%
\centering
        
\begin{tikzpicture}
  [scale=.8,auto=center]
  \node[circle, draw] (n1) at (1,10) {A};
  \node[circle, draw, fill=gray] (n2) at (5,10)  {b};

\draw [->,very thick](n1) -- (n2) ;

\end{tikzpicture}

        \caption{Stability}
        \label{fig:subfig1}
 \end{subfigure}\bigskip
 & 
 \begin{subfigure}[b]{0.45\textwidth}\hfill%
\centering
        
\begin{tikzpicture}
  [scale=.8,auto=center]
  \node [circle, draw](n1) at (1,10) {A};
  \node[circle, draw] (n2) at (1,8)  {C};
\node[circle, draw, fill=gray] (n3) at (5,10)  {b};
\draw [->,very thick](n1) -- (n3)  ;
\draw [->,very thick](n2) -- (n3) ;

\end{tikzpicture}

        \caption{Merging}
        \label{fig:subfig5}
 \end{subfigure}\bigskip\\

 \begin{subfigure}[b]{0.45\textwidth}\hfill%
\centering
   
\begin{tikzpicture}
  [scale=.8,auto=center]
  \node [circle, draw, fill=gray](n1) at (5,10) {b};
 \node (n2) at (1,10){   };
\draw [->,very thick](n2) -- (n1) ;

\end{tikzpicture}

        \caption{Birth}
        \label{fig:subfig3}
 \end{subfigure}\bigskip
 &
 \begin{subfigure}[b]{0.45\textwidth}\hfill%
\centering
      
\begin{tikzpicture}
  [scale=.8,auto=center]
  \node [circle, draw](n1) at (1,10) {A};
  \node [circle, draw, fill=gray](n2) at (5,10)  {b};
\node [circle, draw, fill=gray](n3) at (5,8)  {c};
\draw [->,very thick](n1) -- (n2) ;
\draw [->,very thick](n1) -- (n3) ;

\end{tikzpicture}

        \caption{Splitting}
        \label{fig:subfig4}
 \end{subfigure}\bigskip\\
 
 \begin{subfigure}[b]{0.45\textwidth}\hfill%
\centering
       
\begin{tikzpicture}
  [scale=.8,auto=center]
  \node [circle, draw] (n1) at (1,10) {A};
 
\draw [->,very thick](n1) -- (5,10) ;

\end{tikzpicture}

        \caption{Death}
        \label{fig:subfig2}
 \end{subfigure}\bigskip
 
 & \bigskip\\

\end{tabular}
\end{figure}

Figure \ref{types} presents the five ideal types of knowledge evolution as a proximity network of topics, that we mathematically formalize as follows. Let us consider $M$ topics emerged as the result of a topic modeling exercise from a corpus of articles at time $t$ and $N$ topics at time $t+1$. We tackle the critical problem of tracking the transformation of the set of topics $M=(1,\dots,A,\dots,M)$ at t into the set of topics $N=(1,\dots,a,\dots,N)$ at $t+1$. Specifically, we are interested in measuring the magnitude of the various phenomena such as birth, death, merging, and splitting.
Consider a  similarity index based on word co-occurrence, $simil$\footnote{Typically, this index is the cosine similarity index, as it is used in the empirical part of the paper}, between each couple of topics $(A,a)$ with $A \in M$ and $a \in N$ and consider the similarity matrix S $(M \times N)$ 

\bigskip
\[
\text{S} =\kbordermatrix{
&a&\dots&N& \\
A&simil_{1,1} & \dots. &  simil_{1,N}& \\ 
 \vdots&  & \ddots & &\\ 
M& simil_{M,1} & \dots &  simil_{M,N}& 
 }
\]\\

For the sake of clarity and with reference to Figure \ref{types}, let us consider the minimal example in which $M=(A,B)$ and $N=(a,b)$

\[
  \text{S} =   \kbordermatrix{
    & a & b \\
A& \alpha & \beta \\
B& \gamma & \delta
 }
\]\\

\begin{figure}\caption{bipartite network of topics of two time windows} \label{bim}
\centering
\begin{tikzpicture}
  [scale=.8,auto=left]
  \node [circle, draw](n1) at (1,10) {A};
  \node[circle, draw] (n2) at (3,10)  {B};
\node[circle, draw] (n3) at (5,10)  {C};
\node[circle, draw] (n4) at (7,10)  {D};

\node[circle, draw, fill=gray] (n5) at (1,6)  {a};
\node[circle, draw, fill=gray] (n6) at (4,6)  {b};
\node[circle, draw, fill=gray] (n7) at (7,6)  {c};

\draw [->, thick](n1) -- (n5)  ;

\draw [->, thick](n2) -- (n5)  ;

\draw [->,thick](n3) -- (n7) ;
\draw [->,thick](n3) -- (n6) ;

\end{tikzpicture}
\end{figure}
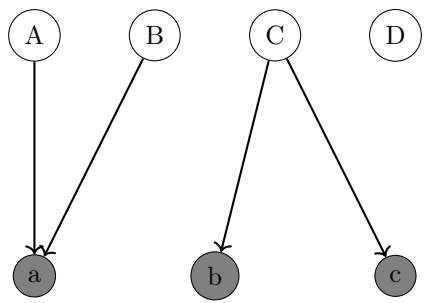

The network representation allows to visualize the five ideal types of knowledge evolution: Table \ref{cases} summarizes them and the necessary and sufficient conditions on the values of the similarity index to observe such cases.
However, with a higher number of topics, a derivation of the conditions on the values of the similarity index would be cumbersome. Moreover, Table \ref{cases} depicts ideal situations only, while the observed reality usually deals with a continuous mixture of the paradigmatic cases presented above. For instance, already in the case with $M=4$ and $N=3$ depicted in Figure \ref{bim}, the analysis becomes strenuous.

\begin{table}
\caption{Bimodal network and empirical indexes}
\label{cases}
\setlength{\tabcolsep}{5mm} 
\def\arraystretch{1.25} 
\centering
\begin{tabular}{|c|c|c|}
  \hline
  Network    &   Matrix param.   &   Cases
  \\ \hline
\begin{tikzpicture}
  [scale=.4,auto=left]
  \node[circle, draw] (n1) at (1,10) {A};
\node[circle, draw] (n2) at (1,8) {B};
  \node[circle, draw, fill=gray] (n3) at (5,10)  {a};
  \node[circle, draw, fill=gray] (n4) at (5,8)  {b};
\draw [->,very thick](n1) -- (n3)  ;
\draw [->,very thick](n2) -- (n4)  ;

\end{tikzpicture}   &   $\alpha , \delta \neq0$  $\gamma, \beta =0$  & STABILITY: no births, no deaths
  \\ \hline
\begin{tikzpicture}
  [scale=.4,auto=left]
  \node[circle, draw] (n1) at (1,10) {A};
\node[circle, draw] (n2) at (1,8) {B};
  \node[circle, draw, fill=gray] (n3) at (5,10)  {a};
  \node[circle, draw, fill=gray] (n4) at (5,8)  {b};

\end{tikzpicture}  &    $\alpha , \delta, \gamma, \beta =0$   & INSTABILITY: births and deaths
  \\ \hline
\begin{tikzpicture}
  [scale=.4,auto=left]
  \node[circle, draw] (n1) at (1,10) {A};
\node[circle, draw] (n2) at (1,8) {B};
  \node[circle, draw, fill=gray] (n3) at (5,10)  {a};
  \node[circle, draw, fill=gray] (n4) at (5,8)  {b};
\draw [->,very thick](n1) -- (n3)  ;
\draw [->,very thick](n2) -- (n3)  ;

\end{tikzpicture}   &  $\alpha , \gamma \neq0$  $\delta, \beta =0$  & MERGING: no deaths, but births
  \\ \hline

\begin{tikzpicture}
  [scale=.4,auto=left]
  \node[circle, draw] (n1) at (1,10) {A};
\node[circle, draw] (n2) at (1,8) {B};
  \node[circle, draw, fill=gray] (n3) at (5,10)  {a};
  \node[circle, draw, fill=gray] (n4) at (5,8)  {b};
\draw [->,very thick](n1) -- (n3)  ;
\draw [->,very thick](n1) -- (n4)  ;

\end{tikzpicture}   &  $\alpha , \beta \neq0$  $\delta, \gamma =0$  & SPLITTING: no births, but deaths
  \\ \hline
  \end{tabular}
\end{table}

With this purpose in mind, we consider the similarity matrix $S$ as the incidence matrix of $M$ over $N$. We can thus employ $S$ to create a bi-adjacency matrix $D$, and thus consider Figure \ref{bim} as the resulting bipartite network in which $M$ and $N$ are the sets of nodes, while the elements of the matrix are the weights of the edges. 

\[
D=
\left[
\begin{array}{c|c}
0 & S \\
\hline
S^T& 0
\end{array}
\right]
= \]
\smallskip
\[=
  \text{}  \kbordermatrix{
    & A & \dots & M &\vrule & a & b & \dots &N\\
    A & 0 & 0 & 0 &\vrule&  &  & &\\
 B & 0 & 0 & 0 &\vrule&  &  & S&\\
    \dots& 0 & 0&0 &\vrule &  &  &&\\
    M & 0 & 0 & 0 &\vrule&  &  &&\\
\hline
    a&  &  &  &\vrule& 0 & 0 & 0&0\\
b&  &  &  &\vrule& 0 & 0 &0&0\\
\dots&  & S^T &  &\vrule& 0 & 0 &0&0\\
    N &  &  &  &\vrule& 0 & 0 &0&0
  }
\]
\bigskip

We now show how this representation can help measure the magnitude of births, deaths, merging and splitting. 

Births and deaths can be easily calculated from the matrix $S$.
A row sum equal to zero highlights a death, while a column sum equals to zero indicates a birth. A death means that the semantic legacy completely disappears while a birth means that a topic carries no semantic similarity with other topics in the past. 
Once again it is important to notice that these cases are extreme scenarios while in the reality we observe a continuum between births and deaths. We might thus calculate an index $Novelty\_i$ ($NI$) for each topic $i$ at time $t+1$ where for $NI\_i=MAX$ we have a birth, that is a topic with no similarity to any other previous one. For higher value we have a higher novelty of the topic. We can also measure an average change in $NI$ on the overall structure of a scientific field by looking at distributions of these indexes over the topics. For instance, let us consider the \textit{Novelty Index} and the average, defining: 

\begin{equation}
NI_{j} = 1-\frac{\sum_{i}^{M}S_{i,j}}{M}  
\end{equation}

where $j$ is the index of the $j$-th column in the matrix S, and

\begin{equation}
NI=1-\frac{\sum_{i}^{M}\sum_{j}^{N}S_{i,j}}{M*N}  
\end{equation}

We take the average of all the cell values in matrix S. If the similarity index is  bounded between 0 and 1, such it is the very common case of the cosine similarity index, thus $NI$ ranges from 0 to 1. For very small value of novelty,  new topics show different word distribution from old ones. 

As mentioned, transformation of topics can take the form of merging and splitting. We say that a merging occurs if a topic at time $t+1$ shows a high similarity with two topics at time $t$, meaning that the semantic universe of $A$ and $B$ at $t$ (as in figure \ref{bim}) is combined in the topic $a$. Similarly, we can say that a split occurs if the semantic legacy of one topic at $t$ is to be found in multiple topics at $t+1$ as in the case for topic $C$. 

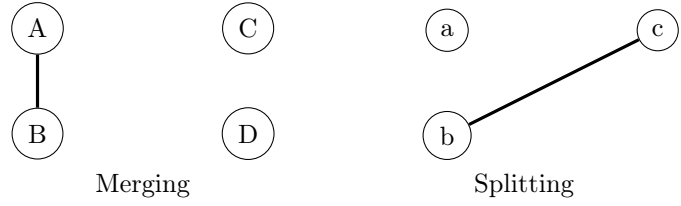
\begin{figure}    \caption{1 mode network, projection of Figure 2}
        \label{merging}
\centering
\begin{tabular}{cccc}
\begin{tikzpicture}[scale=0.7] 

  \node [circle, draw](n1) at (1,10) {A};
  \node[circle, draw] (n2) at (1,8)  {B};
\node[circle, draw] (n3) at (5,10)  {C};
\node[circle, draw] (n4) at (5,8)  {D};
\draw [very thick](n1) -- (n2)  ;
\end{tikzpicture}
&&&
\qquad
\begin{tikzpicture}[scale=0.7]

  \node [circle, draw](n1) at (1,10) {a};
  \node[circle, draw] (n2) at (1,8)  {b};
\node[circle, draw] (n3) at (5,10)  {c};

\draw [very thick](n2) -- (n3) ;

\end{tikzpicture} \\
Merging&&&Splitting
\end{tabular}
\end{figure}

To analyse the intensity of a merging we can project the bipartite network of figure \ref{bim} into its two 1-mode-network of figure \ref{merging}. This is achieved by a matrix multiplication $S\times S^T$ for the merging and $S^T\times S$ for the splitting which result in two matrices $P^{merging}$ and $P^{splitting}$ of dimension respectively $M \times M$ and $N \times N$.
Please note, that for the properties of matrix multiplication  $P^{merging}$ and $P^{splitting}$ are always square matrices, even when the number of topics in two periods differs. 

The network is represented by the matrix P

\[
  \text{$P^{merging}$} =  \kbordermatrix{
    & A & B& ...&M \\
    A &  &  &  \\
 B &  &S\times S^T  & \\
    ...& & & \\
    M &  &  & 
}
\]

\[
  \text{$P^{splitting}$} =  \kbordermatrix{
    & a & b& ...&N \\
    a &  &  &  \\
 b &  &S^T\times S  & \\
    ...& & & \\
    N &  &  & 
}
\]

The matrix transformation allows us to draw the 1-mode-network as in Figure \ref{merging}, which represents the merging and splitting between two time windows. The matrix formulation of the network is also useful for computing the intensity of merging and splitting on the basis of the two relative matrices $P$.
Let us consider the matrix $P^{merging}$ in a minimal example of the table \ref{cases}

\begin{equation}
P^{merging}=S\times S^{T}=\begin{bmatrix}
\alpha  & \beta \\ 
 \gamma & \delta 
\end{bmatrix}\times \begin{bmatrix}
\alpha  & \gamma \\ 
 \beta & \delta 
\end{bmatrix}=\begin{bmatrix}
\alpha \cdot \alpha +\beta \cdot \beta & \alpha \cdot \gamma + \beta \cdot \gamma\\ 
\alpha \cdot \gamma +\beta \cdot \delta & \gamma \cdot \gamma +\delta \cdot \delta
\end{bmatrix}
\end{equation}

The matrix $P$ is always symmetric and, for our purpose, we focus on the low triangle. The merging is captured by the number outside the diagonal  $(\alpha \cdot \gamma +\beta \cdot \delta)$, where $(\alpha \cdot \gamma)$ is the intensity of the merging of $A$ and $B$ in $a$, while $(\beta \cdot \delta)$ is the intensity of the merging of $A$ and $B$ in $b$. In this exemplary case shown in Table \ref{types}, $\beta$ and $\delta$ are equal to zero and $\alpha$ and $\gamma$ are different from zero: thus, we have a merging between $A$ and $B$ as depicted in Figure \ref{merging}.

Mutatis mutandis, we can consider the case of splitting. Once again, the low triangle off the diagonal highlights the intensity of split with $(\alpha \cdot \beta)$ the split of $A$ in $a$ and $b$, while $(\gamma \cdot \delta)$ the split of $B$.

\begin{equation}
P^{split}=S^{T}\times S=\begin{bmatrix}
 \alpha  & \gamma \\ 
 \beta & \delta 
\end{bmatrix}\times \begin{bmatrix}
\alpha  & \beta \\ 
 \gamma & \delta
\end{bmatrix}=\begin{bmatrix}
\alpha \cdot \alpha +\gamma \cdot \gamma & \alpha \cdot \beta + \delta \cdot \gamma\\ 
\alpha \cdot \beta +\gamma \cdot \delta & \beta \cdot \beta +\delta \cdot \delta
\end{bmatrix}
\end{equation}

When we have a large number of topics in both time windows, we can use this formulation to create indexes measuring the intensity of merging and splitting or other properties of the transition.
Specifically, we aim at comparing the values below the diagonal with those on the diagonal. We thus create a normalized matrix in which all elements of the diagonal and below the diagonal add up to one.

\begin{equation}
P^{merging}_{normalized}=P^{merging} \cdot \frac{1}{\sum_{i\leq j}P(i,j)}
\end{equation}
 
In this way, we can compute a $Merging Index$ ($MI$) which takes value $0$ when no merging occurs and it ranges up to an upper limit which can not exceed $1$.
\begin{equation}
MI=1-trace(P_{normalized}^{merging})
\end{equation}

Symmetrically, we calculate a $Splitting Index$ ($SI$) 

\subsection{Conditional dependence}
A last important issue to be addressed consists of the impact of the conditional dependence of topics at time $t$ and its relation with the 1-mode network projection. Two topics at $t$ can appear to merge into a topic at $t+1$ only because they are already similar to each other at time $t$. In this case we might run the risk of identify a spurious process of merging. However, it is possible to account for this dynamic conditional dependence.
We can compute a similarity index among topics at time $t$, $simT$, which can also be represented by a network.

\begin{equation*}
Q = 
\begin{bmatrix}
simT_{1,1} & \dots & simT_{1,M} \\ 
     & \ddots &   \\ 
simT_{M,1} & \dots &  simT_{M,M}
\end{bmatrix}
\end{equation*}\\

Note that $Q$ is a symmetric matrix, with the same dimension $(M \times M)$ as $P^{merging}$. 

\begin{figure}    \caption{Similarity network among topics in t}
        \label{merging1}
\centering
\begin{tikzpicture}
  [scale=.8,auto=left]
  \node [circle, draw](n1) at (1,10) {A};
  \node[circle, draw] (n2) at (1,8)  {B};
\node[circle, draw] (n3) at (5,10)  {C};
\node[circle, draw] (n4) at (5,8)  {D};
\draw [very thick](n1) -- (n2)  ;
\draw [very thick](n2) -- (n3) ;
\draw [very thick](n1) -- (n3) ;
\end{tikzpicture}
\end{figure}
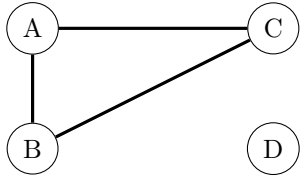

The same procedure can be applied to topics at $t+1$. In this case, we obtain a matrix $(N \times N)$, with the same dimension of $P^{split}$.

In order to take into account the conditional dependence, we might consider $ R^{merging, splitting}= (P^{merging, splitting}|Q^{merging, splitting})$ and recompute the indexes, substituting $R$ with $P$. There exist different ways to operationalize the dependence. Probably the most sophisticated one would be to encode the overall conditional dependence structure within a graphical network \citep{jordan1998,lauritzen1996}. However, we might also consider that the similarity measure has a scalar meaning which goes beyond a simple probabilistic relation. For this reason, we surmise that the conditional dependence can be at best considered by dividing or subtracting element by element the two matrices: in the developed algorithm (see next paragraphs), we divide. Table \ref{final} summarizes the indexes we use and their range.

\begin{table}
\caption{Measuring change in topic modeling}
\label{final}
\centering
\begin{tabular}{|c|c|c|c|}
\hline
Type of change& Index&Min&Max   \\
\hline
Introduction of new semantic areas or legacy from the past& $NI$&0&1 \\
Merging of the semantic content of topics& $MI$&0&1 \\
Splitting of the semantic content of topics& $SI$&0&1 \\
\hline
\end{tabular}
\end{table}

\subsection{The Proposed Algorithm}
\label{sec:algorithm}

This paragraph describes the algorithm which we developed to operationalize the former theoretical approach. 
Our example relies on the Latent Dirichlet Allocation (LDA) \citep{BleiNgJordan2003}, although this methodology does not involve any assumption in the way topics are created. LDA is a generative model that summarizes the documents through a mixture of topics, where each topic is a probability distribution in the dictionary. The algorithm first generates a database which allows query of documents per time period. Thereafter, it divides the dataset into unigrams where stopwords are eliminated according to the NLTK list (\url{www.nltk.org}). Finally, we have applied the \textit{Porter Stemmer} \citep{porter1980algorithm} on individual words. This algorithm transforms (or truncates) every word in a morphological root form. We create a subset per each T time window and compute $N _{t}$ topics using standard LDAs \footnote{\url{https://radimrehurek.com/gensim/}}. On the generated output we are able to compute the three indexes.
For the similarity computation, we use the probability of the first 100 topic's words to generate the vector weights.

\begin{algorithm}[h!t]
\caption{computeSingleWindow(documentSet, $numTopic_t$, $numTopic_{t+1}$)}
\label{algo:singleWin}
\begin{algorithmic}[1]
\STATE $topic_t$ $\leftarrow$ LDA(documentSet, $numTopic_t$)
\STATE $S_t$ $\leftarrow$ computeTopicSimilarity($topic_t$,  $topic_t$)
\STATE $topic_{t+1}$ $\leftarrow$ LDA(documentSet, $numTopic_{t+1}$)
\STATE $S_{t+1}$ $\leftarrow$ computeTopicSimilarity($topic_{t+1}$, $topic_{t+1}$)
\STATE $Q$ $\leftarrow$ computeTopicSimilarity($topic_{t}$, $topic_{t+1}$)
\STATE $R_{merger}$ $\leftarrow$ $S_t*S_t^T$
\STATE $R_{split}$ $\leftarrow$ $S_{t+1}*S_{t+1}^T$
\STATE $Q_{merger}$ $\leftarrow$ $Q*Q^T$
\STATE $Q_{split}$ $\leftarrow$ $Q^T*Q$
\STATE $P_{merger}$ $\leftarrow$ zeros($R_{merger}$.numRow(), $R_{merger}$.numCol())
\STATE $P_{split}$ $\leftarrow$ zeros($P_{split}$.numRow(), $P_{split}$.numCol())
\FOR{i $\leftarrow$ 1..$R_{merger}$.numRow()}
\FOR{j $\leftarrow$ 1..$R_{merger}$.numCol()}
\STATE $P_{merger}[i,j]$ $\leftarrow$ $\frac{R_{merger}[i,j]}{Q_{merger}[i,j]}$
\ENDFOR
\ENDFOR
\FOR{i $\leftarrow$ 1..$R_{split}$.numRow()}
\FOR{j $\leftarrow$ 1..$R_{split}$.numCol()}
\STATE $P_{split}[i,j]$ $\leftarrow$ $\frac{R_{split}[i,j]}{Q_{split}[i,j]}$
\ENDFOR
\ENDFOR
\STATE merger $\leftarrow$ merger(normalize($P_{merger}$))
\STATE split $\leftarrow$ split(normalize($P_{split}$))
\STATE novelty $\leftarrow$ novelty($Q$)
\STATE
\RETURN merger, split, novelty
\end{algorithmic}
\end{algorithm}

\textit{Algorithm \ref{algo:singleWin}} shows the pseudo-code to compute the time window from $t$ to $t+1$. It simply takes in input the cleaned documents of the selected windows and the number of topics at time $t$ and $t+1$ and returns the merging, splitting and value indexes. In details, the algorithm generate a LDA model for each time window $t$ and $t+1$ and computes the similarity between topics at time $t$ and $t+1$ (and themselves). Then, it computes the matrices $P_{merging}$, $P_{splitting}$ using the similarity matrix $S$ and the matrix $Q$. The two $P$ matrices are used to compute \textit{MI} and \textit{SI}, while the matrix $Q$ is used to compute \textit{NI}.

\section{Evaluation}

As to evaluate this approach, we cannot benchmark it with other dynamic methods such as DTM, since we do not track the single topics over time, but we compare adjacent time windows to measure the degree of topics recombination. Therefore, we test the methodology by applying the algorithm  on an artificially-generated dataset with controlled characteristics. 

\subsection{Artificial Data Creation}
\label{sec:toy}
To generate the experimental datasets, we create artificial topics reflecting natural and realistic textual content. Instead of directly producing topics as sets of artificially-built sets of words, we started from \textit{concept seeds}, used as query of real textual data. A concept seed is a word (or compound word) that represents a concept in a text-based resource. For example, the concept seed \textit{physics} within the \textit{Wikipedia} resource is the \textit{Wikipedia} page about \textit{Physics}. 
From a set of concept seeds and their associated Wikipedia pages, it is possible to extract the whole textual content and build artificial documents for the chosen concepts\footnote{We used the  library \textit{Wikipedia} available at \url{https://github.com/goldsmith/Wikipedia}, which acts as a \textit wrapper of the MediaWiki API (\url{https://www.mediawiki.org/wiki/})}.

In the following exercise, we selected 8 concept seeds, all related to the field of Economics, in order to understand how well our approach works on a toy model reflecting contents which are consistent with the real data we used in Section \ref{sec:results}).

As in most natural language processing systems, we applied some pre-processing phase, which includes the removal of stopwords as well as functional linguistic items such as determiners, punctuations, etc\footnote{We used the library \textit{Spacy} (\url{https://spacy.io/}), filtering out the words having the following Part-of-Speech tags: DET (article), NUM (number) and PUNCT (punctuation).}. 

Once the sets of words are built, we generated a document for each seed concept by randomly selecting the words\footnote{The number of words of each document has been chosen randomly.} with uniform probability. We maintained word repetitions to allow us to sampling words with their real frequency and generate documents closed to real cases. The documents generated are  used to train different LDA models with different seeds concepts. 

Finally, we compared the topics of different LDA models by means of the proposed measures to see whether they capture the dynamics of the topic changes. 
We refer the reader to Appendix \ref{app:algo} for details about the  algorithms.

\subsection{Controlled Experiments}
To evaluate the algorithm we create 8 different controlled experiments which are designed to capture the 4 ideal cases of knowledge evolution. Specifically, we conducted twice 4 experiments to test the functioning of the method in 4 different situations by changing (or not) the number of topics and by replacing (or not) the concept seed.
In the first 4 runs we kept the scenario as simple as possible and we slightly increased the complexity of the exercise in the second 4 runs. 

In the former, the number of topics at time $t$ are fixed to 2 for the first experiment and 4 for the second one; the number of topics at time $t+1$ is determined by the experiment (see Table \ref{tab:toy-result2} for details). In details, we set each experiment as follows:

\begin{description}
    \item[stability] the number of topics and seed concepts are kept the same. The variation is only stochastic.
    \item[birth/death] the number of topics  does not change, but we replace the concept seeds to force death of the previous topics and birth of new ones.
    \item[merging] the seed concepts do not change, but we reduce the number of topics to force a situation of merging. For instance, if we cluster the same concept seeds in 2 and 1 topics, we necessarily observe only merging and no splitting. 
    \item[splitting]the seed concepts do not change, but we increase the number of topics to force a situation of splitting. 
\end{description}

\smallskip

Table \ref{tab:toy-result2} summarizes the design of the experiments and depicts avarage values of 100 runs of the \textit{Algorithm \ref{algo:giocattolo}}. Concerning with the first 4 simple designs, experiments are conceived to force the results and create  only splitting and only merging. For the splitting the number of topics increases from one to two and we should not observe merging since at $t-1$ there is also one topic. Analogously, in the case of merging the number of topic shrinks to one in $t+1$.  The remaining two experiments compare stability with births and deaths, which lead to a higher degree of novelty. Our indexes vary as expected: in \textit{splitting} and \textit{merging} the $MI$ and $SI$ respectively go to zero. If we compare \textit{stability} with \textit{births and deaths} the $NI$ is much higher in the former case. Table \ref{tab:toy-result2} shows four different experiments with higher number of topics. It is relevant to notice that even with a few topics, it is impossible to get a clear-cut outcome since the recombination of knowledge may be unexpected and typically reproduces at the same time merging, splitting, stability for some topics, and birth and death for others. However, these baseline examples clearly points at the aggregate behaviour of topics within a discipline.

\begin{landscape}
\begin{table}[h!]
\centering
\caption{The table shows the experimental results conducted over the four cases.}
	\label{tab:toy-result2}
\begin{tabular}{|l|l|l|p{20mm}|p{20mm}|l|l|}
\hline
\textbf{Type of experiment} & \textbf{Concept seed} & \textbf{Replacement} & \textbf{Topics at time t} & \textbf{Topics at time t+1} & \textbf{index}                                                                  & \textbf{value}                                                           \\ \hline
stability           & \begin{tabular}[c]{@{}l@{}}labour economics\\ innovation economics\end{tabular}                       & no                    & 2                     & 2                     & \begin{tabular}[c]{@{}l@{}} $MI$\\ $SI$\\  $NI$\end{tabular} & \begin{tabular}[c]{@{}l@{}}0.189\\ 0.192\\ 0.43\end{tabular}   \\ \hline
splitting               & \begin{tabular}[c]{@{}l@{}}labour economics\\ innovation economics\end{tabular}                       & no                    & 1                      & 2                        & \begin{tabular}[c]{@{}l@{}}$MI$\\ $SI$\\  $NI$\end{tabular} & \begin{tabular}[c]{@{}l@{}}0.0 \\ 0.328 \\  0.155\end{tabular} \\ \hline
merging               & \begin{tabular}[c]{@{}l@{}}labour economics\\ innovation economics\end{tabular}                       & no                    & 2                      & 1                        & \begin{tabular}[c]{@{}l@{}} $MI$\\ $SI$\\  $NI$\end{tabular}  & \begin{tabular}[c]{@{}l@{}}0.398 \\ 0.0\\ 0.296\end{tabular}          \\ \hline
birth/death         & \begin{tabular}[c]{@{}l@{}}labour economics\\ innovation economics\end{tabular}                  & \begin{tabular}[c]{@{}l@{}}cultural economics\\ environmental economics\end{tabular}                    & 2                      & 2                        & \begin{tabular}[c]{@{}l@{}} MI\\ SI\\ NI\end{tabular}  & \begin{tabular}[c]{@{}l@{}}0.363 \\ 0.385 \\ 0.768 \end{tabular} \\ \hline

\\ \hline
stability           & \begin{tabular}[c]{@{}l@{}}labour economics\\ innovation economics\\ cultural economics\\ environmental ecs\end{tabular}                       & no                    & 4                     & 4                     & \begin{tabular}[c]{@{}l@{}}$MI$\\ $SI$\\  $NI$\end{tabular} & \begin{tabular}[c]{@{}l@{}}0.541\\ 0.545\\ 0.555\end{tabular}   \\ \hline
splitting               & \begin{tabular}[c]{@{}l@{}}labour economics\\ innovation economics\\ cultural economics\\ environmental ecs\end{tabular}                       & no                    & 4                      & 8                        & \begin{tabular}[c]{@{}l@{}}$MI$\\ $SI$\\  $NI$\end{tabular} & \begin{tabular}[c]{@{}l@{}}0.574\\ 0.747\\  0.561\end{tabular} \\ \hline
merging               & \begin{tabular}[c]{@{}l@{}}labour economics\\ innovation economics\\ cultural economics\\ environmental economics\end{tabular}                       & no                    & 4                      & 2                        & \begin{tabular}[c]{@{}l@{}}$MI$\\ $SI$\\  $NI$\end{tabular}  & \begin{tabular}[c]{@{}l@{}}0.587 \\ 0.297\\ 0.472\end{tabular}          \\ \hline
birth/death         & \begin{tabular}[c]{@{}l@{}}labour ecs\\ innovation economics\\ cultural economics\\ environmental economics\end{tabular}                  & \begin{tabular}[c]{@{}l@{}}industrial economics\\ transport economics\\ economic history\\ health economics\end{tabular}                    & 4                      & 4                        & \begin{tabular}[c]{@{}l@{}}$MI$\\ $SI$\\  $NI$\end{tabular}  & \begin{tabular}[c]{@{}l@{}} 0.640 \\ 0.662 \\ 0.777 \end{tabular} \\ \hline
\end{tabular}

\end{table}
\end{landscape}

\section{The Evolution of Knowledge in Economics} 
\label{sec:results}

The dataset is a collection of documents which appear in the JSTOR database (\url{www.jstor.org}) and were published from 1845 to 2013 in more than 190 journals concerning with economic sciences (also defined as \textit{economics}). They are more than 460,000 documents, classified as research articles (about 250,000), book reviews (135,000), miscellaneous (73,000), news (4,000) and editorials (500). For each document, in addition to bibliographic information (title, publication date, authors, journal title, etc.), the dataset provides full content in form of a bag of words, i.e. the set of words used in the documents associated with their frequencies.

\begin{figure}[th]
\caption{Distribution of documents in the corpus per year of publication}
\includegraphics[width=1\textwidth]{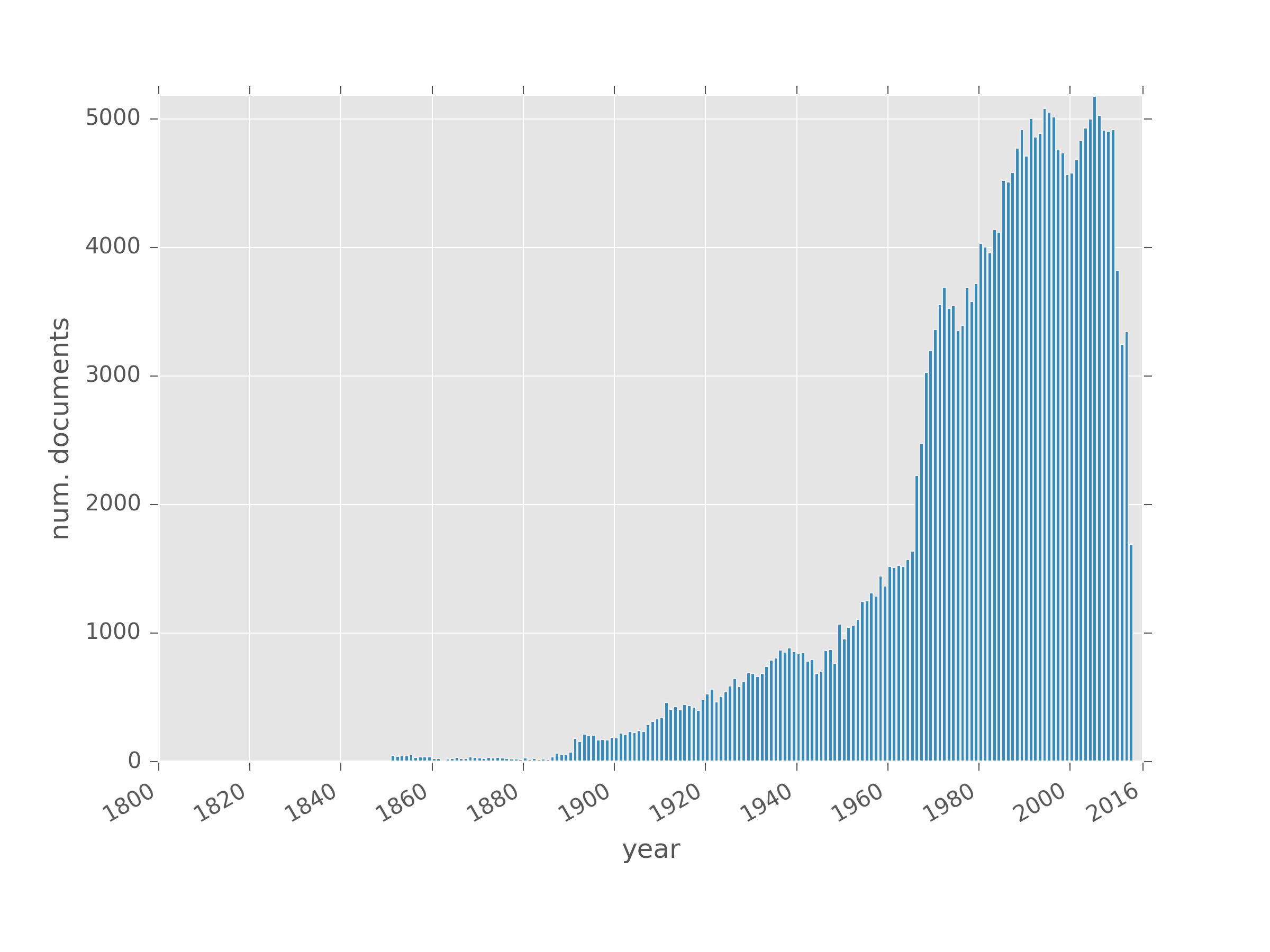}
\label{fig:paper_per_year.png}
\end{figure}

The following analysis only considers the research articles in order to remove the possible noise caused by using different types of documents, which can be written in different languages. The distribution of research articles over the time considered is very skewed (see Figure \ref{fig:paper_per_year.png}). Although the first documents date back to 1845, until the end of the XIX century the corpus of articles accounts only for 2930 items. The increase is almost linear till the beginning of the 1960s, when the number of documents more than doubled in a few years and rose to over 5000 items published every year during the 1990s and 2000s. From 2011 to 2013 we count 8220 items published.

\begin{figure}[h!]
    \centering
    \includegraphics[width=0.7\textwidth]{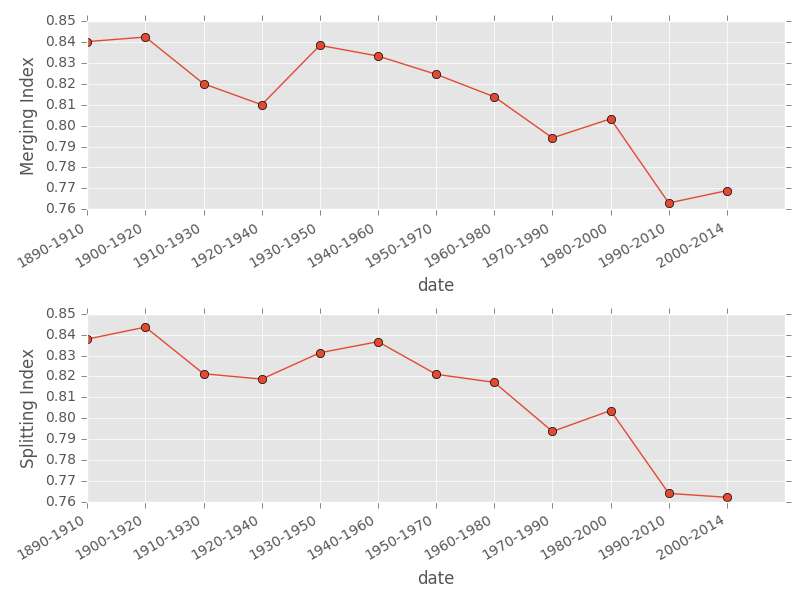}
    \caption{$MI$ and $SI$ - 25 topics 10 years window size}
    \label{fig:25_topic_10_years}
\end{figure}

\begin{figure}[h!]
    \centering
    \includegraphics[width=0.7\textwidth]{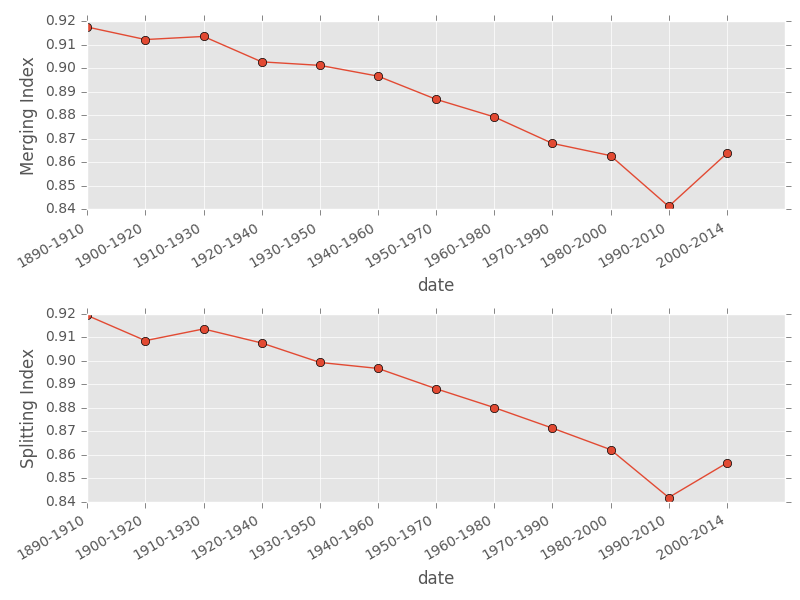}
    \caption{$MI$ and $SI$ - 50 topics and 10 years window size}
    \label{fig:50_topic_10_years}
\end{figure}

\begin{figure}[h!]
    \centering
    \includegraphics[width=0.7\textwidth]{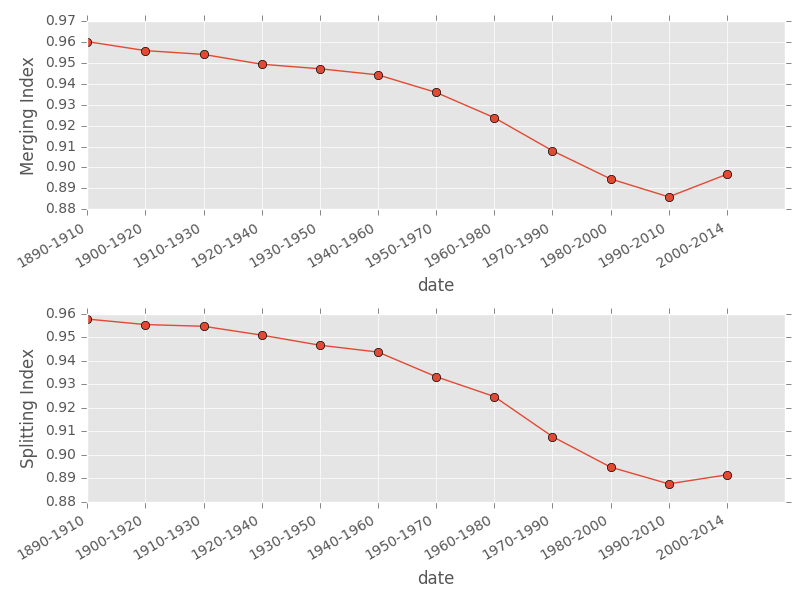}
    \caption{$MI$ and $SI$ - 100 topics and 10 years window size}
    \label{fig:100_topic_10_years}
\end{figure}

\begin{figure}[h!]
    \centering
    \includegraphics[width=0.7\textwidth]{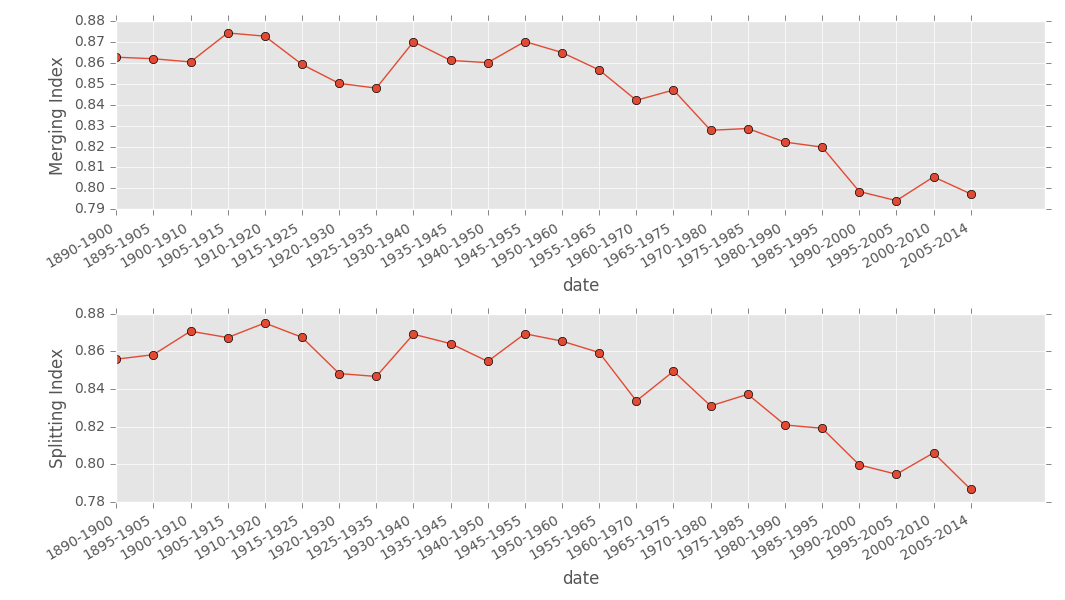}
    \caption{$MI$ and $SI$ - 25 topics and 5 years window size}
    \label{fig:25_topic_5_years}
\end{figure}

\begin{figure}[h!]
    \centering
    \includegraphics[width=0.7\textwidth]{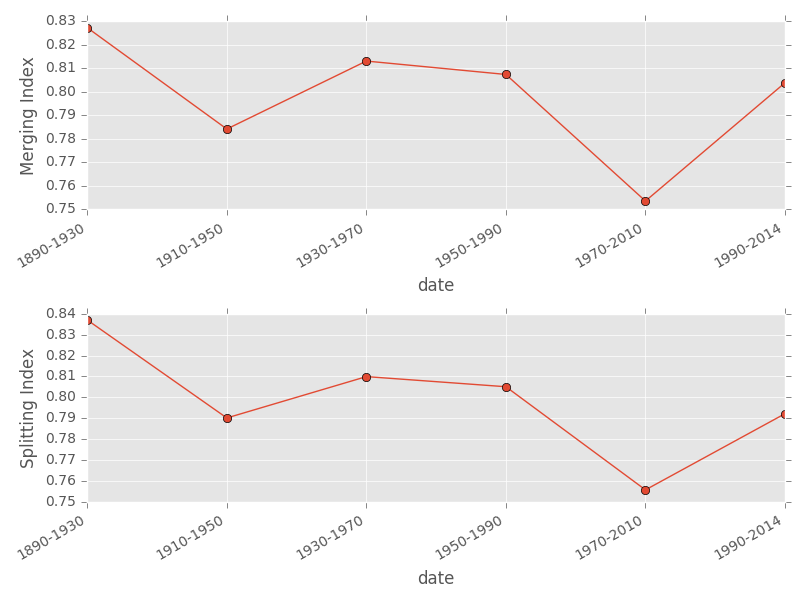}
    \caption{$MI$ and $SI$ - 25 topics and 20 years window size}
    \label{fig:25_topic_20_years}
\end{figure}

The LDA has been applied to research papers published between 1890 and 2013: decades before 1890 were dropped because of the extremely low number of documents. Thereby, the resulting dataset of articles consists of 755,838,336 words and 3,169,515 unique words. 
We experimented varying the hyper-parameters of the method, namely the number of topics and the dimension of time windows, in order to evaluate the robustness and sensitivity of our approach in the 123 years considered. We selected 25, 50 and 100 topics and time windows of 5, 10 and 20 years, keeping fixed one parameter and varying the other one. In details, we first show the values of \textit{SI} and \textit{MI} fixing the window dimension to 10 years and varying the number of topics. In the following figures, for example, 1900-1920 indicates the value of the indexes between 1900 and 1910 compared with the corresponding value between 1910 and 1920.
Figures \ref{fig:25_topic_10_years}, \ref{fig:50_topic_10_years} and \ref{fig:100_topic_10_years} show the indexes for 25, 50 and 100 topics within a window of 10 years. Then, we fixed the number of topics to 25 and we varied the size of the time window. Figures \ref{fig:25_topic_5_years} and \ref{fig:25_topic_20_years} show the indexes for 25 topics and windows of 5 and 20 years. 

These simple tests have demonstrated that the main trends of the indexes do not change substantially by varying the hyper-parameters, meaning that our method is robust to the number of topics and the size of the time windows.

\begin{figure}[h!]
    \centering
    \includegraphics[width=0.7\textwidth]{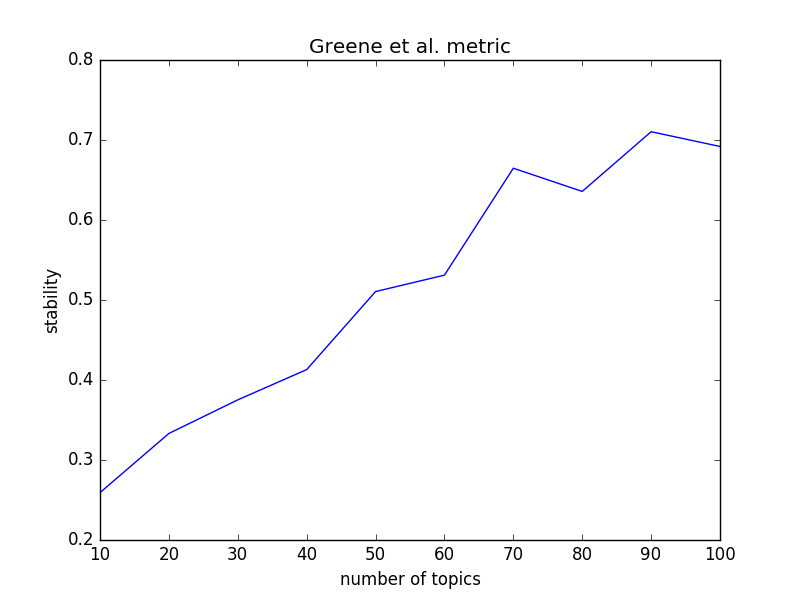}
    \caption{Greene et al.'s stability values for the time window 1910-1920.}
    \label{fig:gr_2}
\end{figure}

\begin{figure}[h!]
    \centering
    \includegraphics[width=0.7\textwidth]{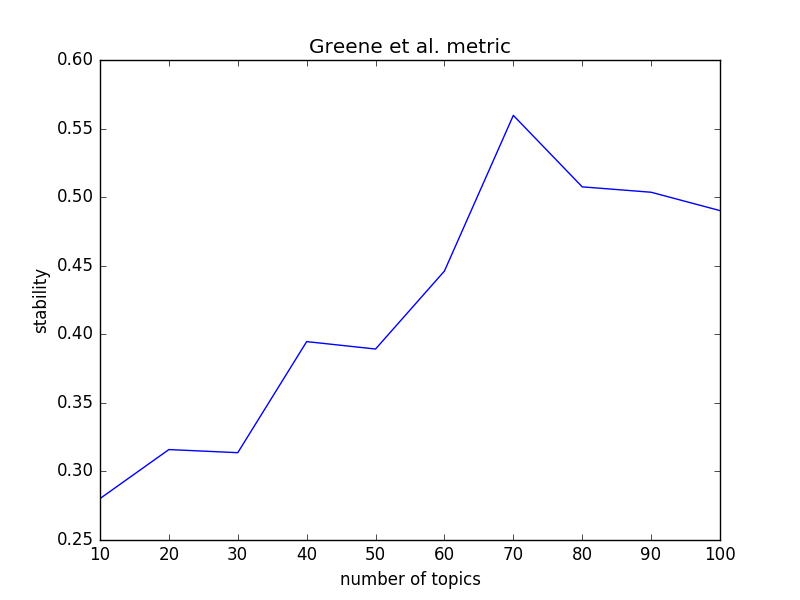}
    \caption{Greene et al.'s stability values for the time window 1940-1950.}
    \label{fig:gr_3}
\end{figure}

To further prove invariance to number of topics and windows-size, we applied Greene metric \cite{Greene2014} on a subset of the research articles with a time windows of 10 years to capture all the possible changing in economic knowledge. Values of the metric reveal how much the topics generated capture the information presented in the dataset. Greene metric requires a range in input, which is formed by the minimum and maximum number of topics, and a step parameter, used by the metric to shift the number of topics considered at the current step starting from the minimum ones. For example, if the minimum number of topics is 10, the maximum is 50 and the step is 20, the Greene metrics will compute a score at 10, 30 and 50 topics. The plot of the metric in Figures \ref{fig:gr_2} and \ref{fig:gr_3} concerns with two windows and shows that increasing the number of topics we can increase stability too, but of course, it becomes very difficult to interpret the meaning of each topic.

As suggested by \citet{mimno2011} when topic modeling is employed to explore the content of a dataset -as in this paper - rather than to predict there is not a definitive test to support the choice of the optimal number of topics. We solved this trade-off between stability and meaningfulness by manually controlling for the topics generated by the model with 25 topics within time window of 10 years. When we found that a few topics could be split up again because they were too general, we set an optimal and analytically useful number of topics to 27. 
Therefore, the following analysis is based on 27 topics within time windows of 10 years, which perform the maximum stability of the indexes varying the number of topics.



\begin{figure}[h!]
\centering
\caption {$MI$ and $SI$ - 27 topics 10 years}  
\includegraphics[width=0.7\textwidth]{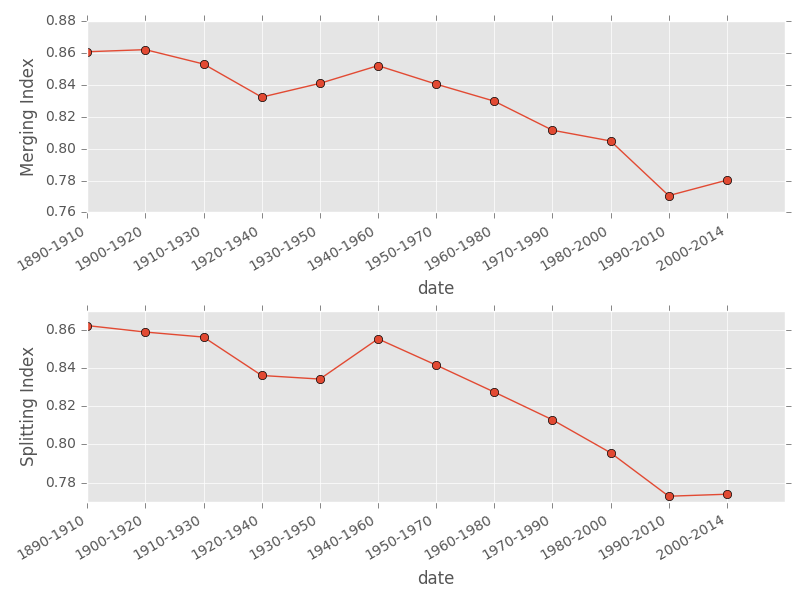}
\label{fig:merger_graph.png}
\end{figure}

\begin{figure}[h!]
\centering
\caption {$NI$ - 27 topics 10 years}
\includegraphics[width=0.7\textwidth]{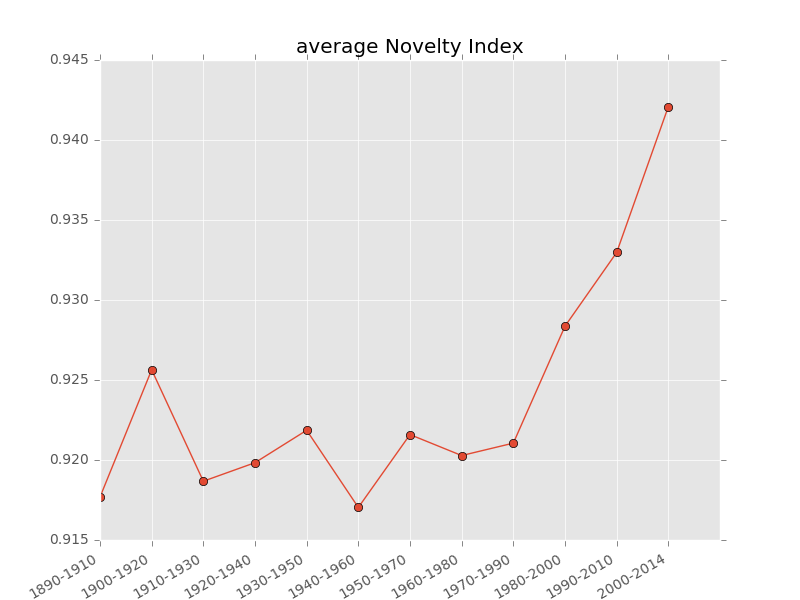}
\label{avgnovelty_full.png}
\end{figure}

Figure \ref{fig:merger_graph.png} shows the values of \textit{MI} and \textit{SI} respectively for each time window, as defined in Section \ref{sec:algorithm}. In the corpus we analyzed, both indexes show a general trend of decreasing values over time, which becomes particularly severe starting from 1960s. Merging and splitting increase only between the 1940s and the 1950s while dropping dramatically in the second half of the XX century. The transformation of topics seems to find new urge only around the end of the century, when merging is increasing againg and splitting is stable.   
As for the $NI$, we mentioned that the index tends to one when new topics emerge without matching with topics at $t-1$. On the average the value is higher than 0.9 all over the 123 years considered, so we tracked both micro-variation and general trend.  In Figure \ref{avgnovelty_full.png} $NI$ does not show relevant variations until 1990s, with some local maximum in the first decade of the past century and a local minimum around the half of it. In the last decade of the century it  grows sharply, revealing a higher rate of brand new topics or at least of topics defined by new words. 

Such a methodological approach has the advantage of tracking the evolution of each single stream of economic theory by looking simultaneously at all the others. On the whole, the analysis of such a big corpus of documents suggests that merging and splitting cannot be considered as opposite phenomena, but a complementary measure of recombination of topics. In particular, trends in the field of economics suggest a steady decrease of both splitting and merging only temporally balanced by a weak growth before and after the WWII. From a historical perspective this is absolutely consistent with the need of theoretical elaboration in economics following the great Depression in 1929 and the dramatic economic changed imposed by the post war reconstruction. 
During the 1960s and in combination with the boom of academic publications, many topics are spreads over a relevant number of documents and journals, although they seem to elaborate on relative stable basis of autonomous topics. Only by the end of the century we have witnessed the development of new-brand topics. The birth of new topics strengthens the hypothesis of self-standing topics shaped by their own specialised language and a lesser exchange of knowledge across the economic discipline. In other words, the terrific expansion of the academic production seems to come with a fragmentation and dispersion in multiple niches of knowledge \cite{cedrini2017} which elaborate on a new language, but not necessarily producing new paradigms. 

\begin{figure}[h!]
\centering
\caption{Combined graph of $SI$ and $NI$}
\includegraphics[width=0.85\textwidth]{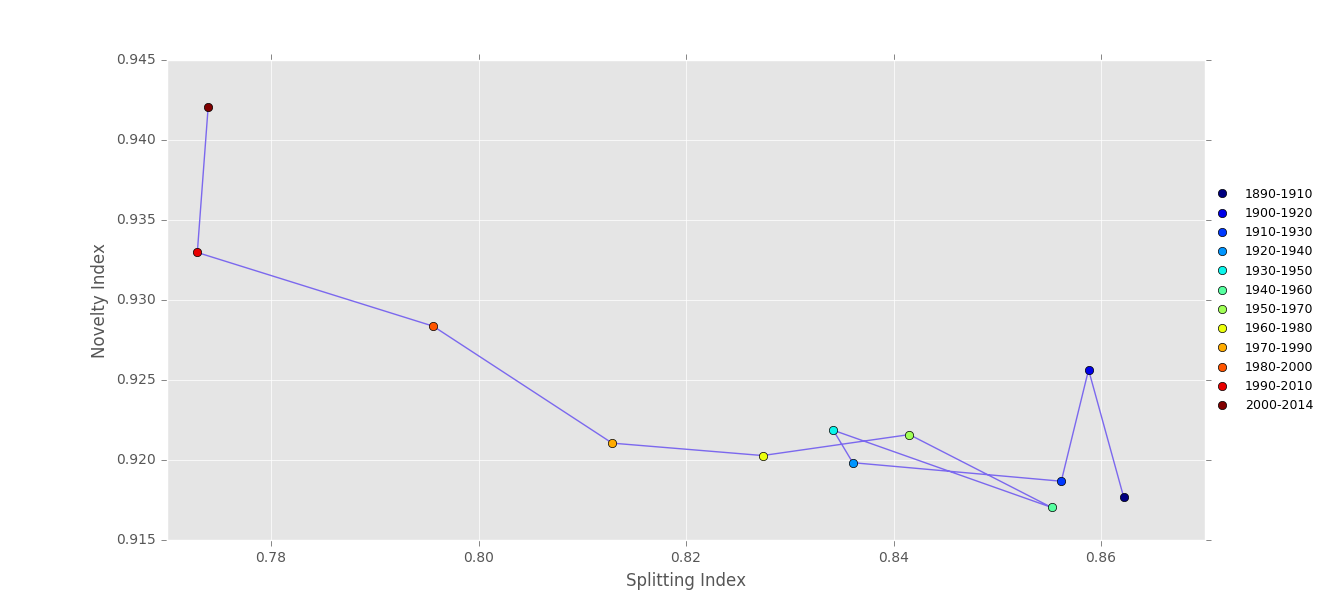}
\end{figure}

\begin{figure}[h!]
\centering
\caption {Combined graph of $MI$ and $NI$}
\includegraphics[width=0.8\textwidth]{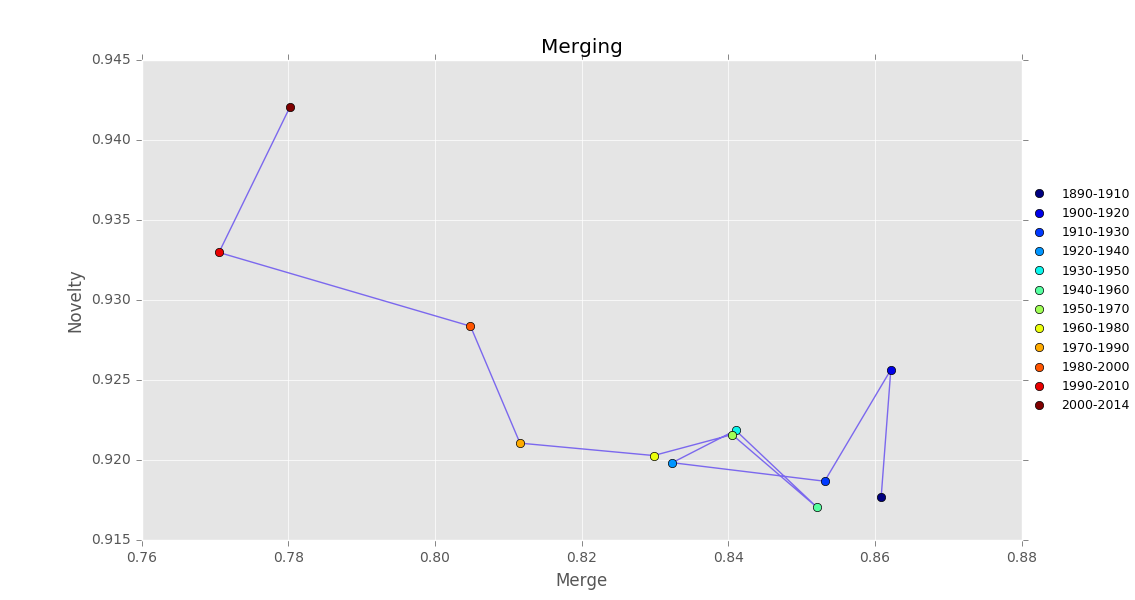}
\end{figure}

\section{Conclusion}
In this paper we proposed a method to measure the evolution of knowledge in a scientific field extracting topics in a corpus of documents. Topic modeling techniques are becoming increasingly refined in treating large and complex corpora of documents, but they may lack of a theoretical reflection of the underlying empirical phenomenon.
Taking a dynamic perspective we recognise five paradigmatic cases of knowledge evolution. We then surmise that modeling the proximity between topics of different time windows as a proximity network might be a useful tool to measure their knowledge dynamics. Indeed, this network approach allows us to develop 3 indexes, which grasp i) the stability of topics over time measuring their rate of death and birth (\textit{Novelty Index} - $NI$), and ii) the degree of recombination of topics (\textit{Merging Index - $MI$} and \textit{Splitting Index - $SI$}). For very simple cases, we are also able to analytically derive those conditions, which link the proximity network and the value of each index. 
 Testing the algorithm over a set of simulated documents, we showed its robustness for each the indexed developed. Finally, we applied our approach to a real and large corpus of academic publications in economics to illustrate how the combined use of $MI$, $SI$ and $NI$ is effective to understand dynamics and trends in economic knowledge and thought.
We believe, this is a first step towards the development of a closer connection between algorithms for dynamic topic modeling and the empirical phenomenon they are supposed to describe.

\appendix

\section{Artificial Data Creation: Algorithms}
\label{app:algo}

In \textit{Algorithm \ref{algo:giocattolo}}, the function \textit{getNum(minNum,maxNum)} returns a number, randomly selected, between \textit{minNum} e \textit{maxNum}; the \textit{getWord()} function returns a word, randomly chosen on the selected set; the function \textit{computeTopicSimilarity()} calculates the cosine similarity between the input topics; the function \textit{zeros()} returns an array containing all zeros.
Finally, the function \textit{getWordList(concept)} generates a set of words. The words are taken from the Wikipedia page that points to the chosen concept. 

In rows [1-6], the function \textit{getWordList} collects, for each concept seed, a set of words. In details, \textit{getWordList}, as shown in \textit{Algorithm \ref{algo:words}}, extracts all words contained both in the Wikipedia page related to the concept in input 
through the python library \textit{Wikipedia}\footnote{\url{https://github.com/goldsmith/Wikipedia}}. Words are extracted using the library \textit{Spacy}\footnote{\url{https://spacy.io/}} and stored in \textit{wordList}\footnote{There exists a \textit{wordList} for each \textit{conceptSeed} in input.}. Then, the \textit{wordList} of each \textit{concept seed} is inserted into \textit{wordConceptList}.
In rows [7-16], \textit{Algorithm \ref{algo:giocattolo}} generates a document for each concept, sampling words (with uniform probability) from the \textit{wordList} related to the \textit{concept seed}. The number of words to sample is specified by \textit{numWords}, which ranges from 1000 to 10000. Successively, in rows [18-20], the algorithm divides documents in two sets, a set containing the first \textit{numDocument} documents and a set containing the remain documents, and applies LDA. The LDA can be applied over the two documents sets or only over a single documents set according to the \textit{replaceDoc} flag. If \textit{replaceDoc} is set to \textit{True}, the first documents set is replaced with the second one (it is set to \textit{False} by default). 

\textit{Algorithm \ref{algo:wordList}} shows how words are processed. We filtered stopwords and words having \textit{Part-Of-Speech} tags \textit{Det} (Determiner), \textit{X} (foreign word), \textit{NUM} (Numeral), \textit{Punct} (Punctuation), \textit{SPACE} and \textit{EOL} (end of line symbols). We also filtered words that does not match the python regular expression \textit{$\backslash$w+}. Furthermore, all unfiltered words are brought back to their morphological root.

\begin{algorithm}[h!t]
\caption{ToyEvaluation(seedConcepts, numDocument, $numTopic_t$, $numTopic_{t+1}$, replaceDoc)}
\label{algo:giocattolo}
\begin{algorithmic}[1]
\STATE wordsConceptList = \{\}
\STATE // create a words list for each concept seed
\FOR{concept in seedConcepts}
\STATE wordsList $\leftarrow$ getWordList(concept)
\STATE wordsConceptList.append(wordsList)
\ENDFOR
\STATE documents = \{\}
\FOR{i $\leftarrow$ 1..len(seedConcepts)}
\STATE numWords $\leftarrow$ getNum(1000, 10000)
\STATE document = \{\}
\FOR{j $\leftarrow$ 1..numWords}
\STATE word $\leftarrow$ wordsConceptList[i].getWord()
\STATE document.append(word)
\ENDFOR
\STATE documents.append(document)
\ENDFOR
\STATE // get topic
\STATE documentSet $\leftarrow$ documents[1:numDocument]
\STATE $topic_t$ $\leftarrow$ LDA(documentSet, $numTopic_t$)
\STATE $M_t$ $\leftarrow$ computeTopicSimilarity($topic_t$,  $topic_t$)
\IF{replaceDoc $\not$= False}
\STATE documentSet $\leftarrow$ documents[numDocument:len(seedConcepts)]
\ENDIF
\STATE $topic_{t+1}$ $\leftarrow$ LDA(documentSet, $numTopic_{t+1}$)
\STATE $M_{t+1}$ $\leftarrow$ computeTopicSimilarity($topic_{t+1}$, $topic_{t+1}$)
\STATE
\STATE /* it then continues as computeSingleWindow algorithm */
\end{algorithmic}
\end{algorithm}

\begin{algorithm}[h!t]
\caption{getWordList(concept)}
\label{algo:words}
\begin{algorithmic}[1]
\STATE posTags $\leftarrow$ \{X, NUM, DET, PUNCT\}
\STATE parser $\leftarrow$ parser(lan=eng)
\STATE wordList $\leftarrow$ \{\}
\STATE wordList $\leftarrow$ getWordList(content, posTags)
\RETURN wordList
\end{algorithmic}
\end{algorithm}

\begin{algorithm}[h!t]
\caption{getWords(content, posTags)}
\label{algo:wordList}
\begin{algorithmic}[1]
\STATE words $\leftarrow$ \{\}
\STATE wikiPage $\leftarrow$ Wikipedia.getPage(concept)
\FOR{sentence in parser(wikiPage.content).sentences}
\FOR{word in sentence.words}
\IF{$\lnot$(word in stopwords) $\land$ $\lnot$(word.pos in posTags) $\land$ match(word,$\backslash$w+)}
\STATE{words.append(word.lemma)}
\ENDIF
\ENDFOR
\ENDFOR
\RETURN words
\end{algorithmic}
\end{algorithm}

\clearpage

\bibliographystyle{abbrvnat}
\bibliography{biblio}

 \end{document}